# Large-Scale Cargo Distribution

Luka Stopar, PhD
Researcher
Jozef Stefan Institute
Jamova cesta 39
1000 Ljubljana, Slovenija
luka.stopar@ijs.si

Luka Bradesko, PhD
Researcher
Jozef Stefan Institute
Jamova cesta 39
1000 Ljubljana, Slovenija
luka.bradesko@ijs.si

Tobias Jacobs, PhD
Senior Researcher
NEC Laboratories Europe GmbH
Kurfürsten-Anlage 36
69115 Heidelberg
tobias.jacobs@neclab.eu

Azur Kurbašić
Researcher
Jozef Stefan Institute
Jamova cesta 39
1000 Ljubljana, Slovenija
azurkurbasic@gmail.com

Miha Cimperman, PhD
Researcher
Jozef Stefan Institute
Jamova cesta 39
1000 Ljubljana, Slovenija
miha.cimperman@ijs.si

## ABSTRACT

This study focuses on the design and development of methods for generating cargo distribution plans for large-scale logistics networks. It uses data from three large logistics operators while focusing on cross border logistics operations using one large graph.

The approach uses a three-step methodology to first represent the logistic infrastructure as a graph, then partition the graph into smaller size regions, and finally generate cargo distribution plans for each individual region. The initial graph representation has been extracted from regional graphs by spectral clustering and is then further used for computing the distribution plan.

The approach introduces methods for each of the modelling steps. The proposed approach on using regionalization of large logistics infrastructure for generating partial plans, enables scaling to thousands of drop-off locations. Results also show that the proposed approach scales better than the state-of-the-art, while preserving the quality of the solution.

Our methodology is suited to address the main challenge in transforming rigid large logistics infrastructure into dynamic, just-in-time, and point-to-point delivery-oriented logistics operations.

## 1. INTRODUCTION

The complexity of operations in the logistics sector is growing, so is the level of digitalization of the industry. With data driven logistics, dynamic optimization of basic logistics processes is at the forefront of the next generation of logistics services.

Finding optimal routes for vehicles is a problem which has been studied for many decades from a theoretical and practical point of view: see [2] for a survey. The most prominent case is the Traveling Salesperson Problem (TSP), where the shortest route for visiting $n$ locations using a single vehicle has to be determined. What is typically associated with the Vehicle Routing Problem (VRP) is a generalization of TSP where multiple vehicles are available. This class of routing problems is notoriously hard; it not only falls into the class of NP-complete problems, but also in practice it cannot be solved optimally even for moderate instance sizes.

Nevertheless, due to its practical importance, many heuristics and approximation algorithms for the vehicle routing problem have been proposed. Bertsimas et al. propose to an integer programming based formulation of the Taxi routing problem and present a heuristic based on a max-flow formulation, applied in a framework which allows to serve 25,000 customers per hour. A heuristic based on neighborhood search has been presented by Kytöjoki et al. in [4] and evaluated on instances with up to 20,000 customers. A large number of natural-inspired optimization methods have been applied to VRP, including genetic algorithms [7], particle swarm optimization [8], and honey bees mating optimization [9].
The particular approach of partitioning the input graph for VRP has been proposed by Ruhan et al. [5]. Here k-means clustering is combined with a re-balancing algorithm to obtain areas with balanced number of customers. Bent et al. study the benefits and limitations of vehicle and customer based decomposition schemes [6], demonstrating better performance with the latter.

In this paper, we present a methodology for large-scale parcel distribution, by utilizing optimization methods with large graph clustering. The paper is structured as follows. In Section 2, we present the technical details of the proposed methodology. We explain the algorithms and data structures used in each of the steps and discuss the interfaces required to link the steps into a working system. In Section 3, we demonstrate the performance of our methodology on two real-world use cases and compare it to the state-of-the-art on synthetic datasets. Finally, in Section 4 we include key findings, summarizing the strengths and limitations of the proposed approach.

## 2. METHODOLOGY
## 2.1 Overview

In this section, we present the details of the proposed methodology for large-scale cargo distribution planning. The methodology, illustrated in Figure 1, uses a three-step, divide and conquer approach to cargo distribution, where we reduce the size of the optimization problem by (i) abstracting the physical infrastructure into a sparse graph representation, (ii) partitioning the graph into smaller chunks (i.e. regions) and (iii) planning the distribution in each region independently. This allows us to run the optimization on large graphs while producing better local results.

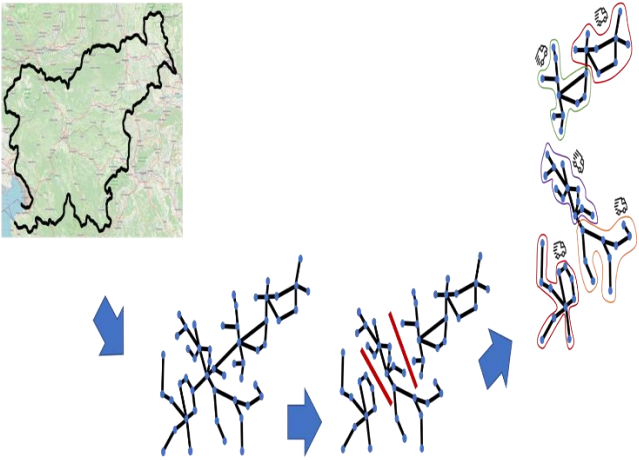

**Figure 1: Three step methodology for logistics optimization.**

Initially, we create a representation of the physical infrastructure as an abstract graph, representing each pickup and drop-off location as a node with edges as shortest connections on road in between.

Next, we partition the abstract graph with a spectral partitioning approach. The method is an adaptation of [10] to graphs, where we use the first $k$ eigenvalues and eigenvectors of the graphs' Laplacian to construct the partitions. In each partition, we construct a distribution plan using an iterative search algorithm. From an initial solution, the algorithm constructs a linear search path by changing the position of a node in the distribution plan. To avoid local minima, it uses design-time blacklist rules which prevent the algorithm from oscillating in a local neighborhood. Each step is described in more details in the following sections.

## 2.2 Graph Construction

For graph construction, the Dijkstra SPF algorithm [11] was applied to identify neighbor relationships between the nodes in the OpenStreetMaps (OSM) dataset and construct the graph representation. By mapping post offices to the closest node on OSM, we tag the post office nodes for SPF search.

The search frontier is a baseline for the SPF procedure and represents the list of nodes whose graph neighbors are to be searched. The final graph is built by iterating with the SPF procedure through the list of all post offices in physical infrastructure (graph nodes), and consolidating results into final the sparse matrix – each iteration computes one row of the matrix.

## 2.3 Graph Partitioning

The partitioning step first represents the graph as a transition rate matrix $(Q)_{ij} = q_{ij}$, where $q_{ij}$ represents the rate of going from node $i$ to node $j$ and is computed as the inverse minimal travel time (obtained from step 1) between the two nodes. With this approach, the rate of going from $i$ to $j$ is represented in terms of the number of possible trips that the driver can make between the two locations in one hour.

The algorithm works by approximating the minimal $k$-cut of the graph, removing its edges and thus reducing the graph to $k$ disconnected components. We adapt a spectral partitioning algorithm introduced in [10] to graphs.

The algorithm first symmetrizes the transition rate matrix as $Q_s = \frac{1}{2}(Q + Q^T)$, to ensure real-valued eigenvalues, and computes its Laplacian:

$$L = I - diag(Q_s \vec{1})^{-1} Q_s$$

Next, it computes the $k$ eigenvectors of $L$, corresponding to the smallest $k$ eigenvalues. It then discards the eigenvector corresponding to $\lambda_1 = 0$ and assembles eigenvectors $v_2, v_3, \ldots, v_k$ corresponding to eigenvalues $\lambda_2 \leq \lambda_3 \leq \cdots \leq \lambda_k$ as columns of matrix $V$. The rows of $V$ are then normalized and used as input to the *k-means* clustering algorithm which constructs the final partitions.

## 2.4 Vehicle Routing

The vehicle routing step uses *Tabu search* [12] to construct the distribution plan. Starting with an initial solution, *Tabu search* constructs a linear search path by iteratively improving the solution in a greedy fashion until a stopping criterion is met. To avoid converging to local minima, *Tabu search* blacklists recent moves and/or solutions for one or more iterations using design-time rules.

In each iteration, the search process generates new possible solutions by removing a node from its current route and placing it after one of the other nodes in the graph, possibly on a different route. To mitigate scaling problems associated with generating $O(n^2)$ possible moves in each step, the algorithm only considers a handful of moves. Specifically, the probability of considering placing node $i$ after node $j$ is proportional to the inverse of the Euclidean distance $d(i,j)$ between the nodes.

Like other local search algorithms, *Tabu search* starts from an initial feasible solution which is constructed using a construction-based heuristic algorithm. The heuristic procedure iteratively selects a node and places it after one of the other nodes in a way that minimizes the travel distance. The procedure iterates until all values are initialized.

## 3. DEMONSTRATION AND RESULTS

In this section, we demonstrate the effectiveness of the proposed methodology on two real-world use cases and compare the methodology to the state-of-the-art in vehicle routing. The first pilot included two national logistics operators, namely Hrvatska Posta (Croatia) and Posta Slovenije (Slovenia). As the main focus of future logistics in Europe is to operate as one large homogenous logistics infrastructure, the two infrastructures were considered as one logistics graph. The second pilot included Hellenic Post (Greece) graph representation and data.

In initial testing, simulated data were used for modelling parcel flow with graph abstraction, graph processing, and optimization responses. The final instances were constructed from real infrastructure data to test the functionalities. The results are presented in the following subsections.

## 3.1 Evaluation on Large Synthetic Graphs

We now demonstrate the scalability of the proposed methodology by comparing its performance to the performance of the baseline *Tabu search* algorithm on synthetic graphs of various sizes, comparing both algorithms' running time and the total travel time in the generated cargo distribution plan. Our results show that the proposed methodology enables fast generation of distribution plans on graphs of up to 10,000 nodes, while also improving the quality of the generated result.

We simulate the logistics infrastructure by generating random planar graphs representing the road network and drop-off locations. First, we generate a cluster of $n$ drop-off locations by sampling a Gaussian distribution around $k$ randomly chosen locations. Next, we connect the locations with Delaunay triangulation [13], resulting in a planar graph. We compute the distance between two locations using the Euclidean metric and assign a 50 $km/h$ speed limit to intra-city edges and a 90 $km/h$ speed limit to inter-city edges. Part of a synthetic graph with 10,000 nodes is shown in Figure **2** below.

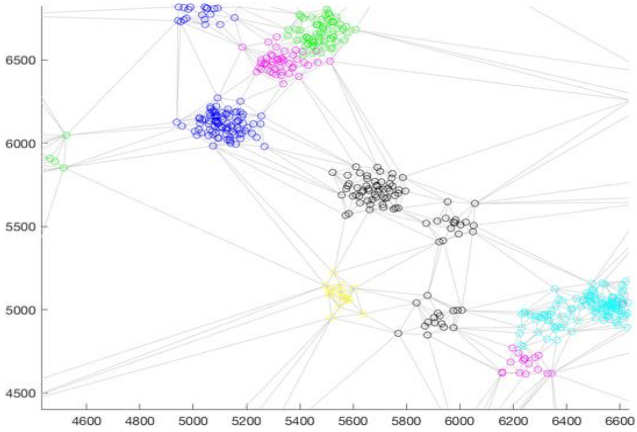

**Figure** 2: Representation of simulated graph with 10,000 nodes.

Table **1** summarizes the computation times of the proposed method along with the quality of the generated distribution plan and compares the results to *Tabu search* without prior clustering. We measure the quality of the generated distribution plan as the distance travelled by all vehicles according to the plan. In each row, we show the average of 10 trials on 10 different graphs.

**Table** 1**: Comparison of efficiency of Tabu search and proposed methodology.**

| Graph Size | Proposed Methodology | | Tabu search | |
|---|---|---|---|---|
| | **Running Time** | **Travel Distance [km]** | **Running Time** | **Travel Distance [km]** |
| **1000** | 6.07min | 64.7k | 0.76min | 85.5k |
| **2000** | 10.07min | 122.9k | 2.98min | 160.8k |
| **5000** | 30.14min | 259.2k | 60.04min | 428.2k |
| **7000** | 39.29min | 377.9k | 166.79min | 577.1k |
| **10000** | 55.64min | 552.2k | 10.78h | 845.1k |

For the experiments we used a *Tabu* list with a length of 5% of the entities (locations) that the algorithm must check, and terminated the algorithm when there was no improvement in the solution for more than 10 seconds.

On large graphs, we see that the proposed methodology significantly reduces the computation time while preserving the quality of the result. The proposed methodology reduces the computation time on graphs larger than 5k nodes, providing a substantial saving of 91% on graphs with 10k nodes. We also observe that the quality of the output slightly improved when applying our divide-and-conquer methodology over *Tabu search*. The improvement ranges between 23% and 40% and is largely attributed to the significantly reduced search space in the partitions as compared to the entire graph.

## 3.2 Testing the instances on pilot use cases

The methods presented and tested on synthetic graphs were also tested on data from two pilot scenarios, namely Slovenian-Croatian post (Pošta Slovenije & Hrvatska Pošta) and Hellenic Post (Greece). In the pilot use cases, the analytical pipeline is used to process ad-hoc events in the logistics infrastructure. The ad-hoc events included were structured into three categories: new parcel request (ad-hoc order), event on distribution objects (vehicle break down) and events related to changes in border crossings – border closed (cross border event).

The instances built on simulated data were loaded with OpenStreetMaps data for abstraction of real infrastructure description into graph representation, as illustrated in Figure 4.

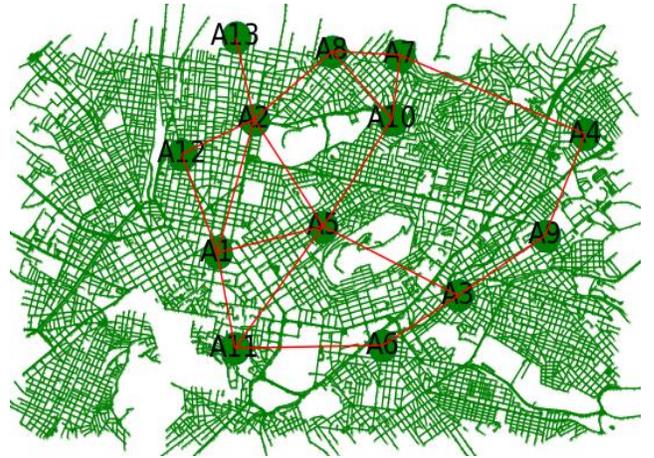

**Figure** 4**: A region of Posta Slovenia graph representation, using OpenStreetMap.**

A similar approach was used for the case of Hellenic Post, where the OSM data for the region of Greece were loaded into the graph abstraction instance. For traffic modelling of the vehicles, the SUMO simulator [14] was used with the regional map. For graph manipulations, the SIoT infrastructure was used to generate the social graph when an ad-hoc event was triggered. The social graph represented all entities (vehicles, etc.) in the infrastructure that are in the scope to be included in event processing. In this way, distribution objects were mapped to physical infrastructure for loading the objects into the graph representation for further optimization and distribution plan estimation

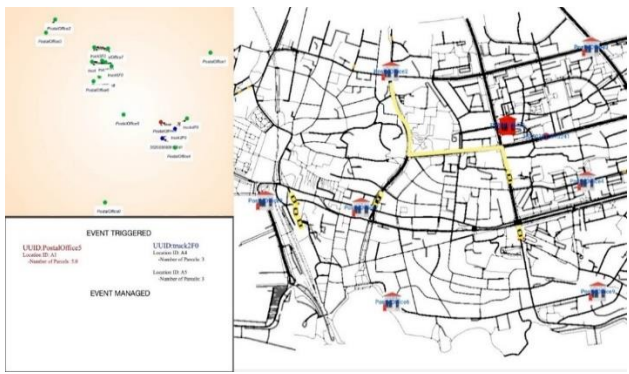

**Figure** 4**: Processing ad-hoc order on a pilot scenario, using SUMO simulator.**

An example of the social graph generation and ad-hoc event processing is presented in Figure 4, where a new ad-hoc request is processed by SIoT and analytical pipeline.

The results show that abstracting the logistics infrastructure and clustering the graph into regional structures enabled real-time processing of complex events in the logistics infrastructure. The response time for processing an ad-hoc event in regions of between 50 and 100 nodes was between 20 and 30 seconds. This is relatively fast compared to alternatively processing 1000 nodes or more

## 4. CONCLUSION

In this paper, we presented an approach for generating cargo distribution plans on large logistic infrastructures. Our results show that the proposed approach can scale to graphs of up to 10,000 nodes in practical time while preserving and even slightly improving the quality of the result.

Since the main use case of logistics is point-to-point regional delivery and just-in-time delivery, these new services are oriented exactly to regional logistics optimization. More importantly, the approach enables to process ad-hoc events, such as new parcel delivery requests, events related to distribution vehicles, or to infrastructure. The ad-hoc event processing includes manipulating the graph representation and running the optimization methods in real-time. Since our method clusters and regionalizes large graphs, such approach can enable real-time processing of events on large graphs, by limiting the changes to the affected regional parts of the infrastructure.

However, while our approach can be combined with several state-of-the-art methods, its main drawback remains the inability to generate inter-region routes, making it suitable only for local and last-mile distribution plans. Future work will focus on investigating the generation of inter-region plans and connecting multiple regions into one distribution plan. Some of the options include introducing border checkpoints where cargo can be handed over to vehicles of neighboring regions, using dedicated inter-region "highway" channels, and using dedicated vehicles for cross-region deliveries.

## 5. ACKNOWLEDGEMENTS

This paper is supported by European Union's Horizon 2020 research and innovation programme under grant agreement No 769141, project COG-LO (COGnitive Logistics Operations through secure, dynamic and ad-hoc collaborative networks).